\documentclass{article}
\usepackage{iclr2027_conference,times}
\usepackage{booktabs}
\usepackage{graphicx}
\usepackage{makecell}
\usepackage[table]{xcolor}
\usepackage{amsmath}
\usepackage{amssymb}
\usepackage{booktabs}
\usepackage{graphicx}
\usepackage{microtype}
\usepackage{multirow}
\usepackage{float}
\usepackage{tabularx}
\usepackage{tikz}
\usepackage{xcolor}
\usepackage{hyperref}
\usepackage{algorithm}
\usepackage{algorithmic}
\usepackage{amsmath}

\usepackage{url}
\usepackage{xspace}
\usepackage{duckuments}
\usepackage{wrapfig}
\usepackage{subcaption}
\usepackage{caption}
\renewcommand{\arraystretch}{1.08}
\usepackage{xcolor}
\usepackage{colortbl}
\usepackage{listings}

\definecolor{lightgray}{gray}{0.93}

\usetikzlibrary{arrows.meta,positioning}

\definecolor{edgeblue}{HTML}{2563EB}
\definecolor{cloudpurple}{HTML}{7C3AED}
\definecolor{rewardgreen}{HTML}{16803A}
\definecolor{lightblue}{HTML}{EFF6FF}
\definecolor{lightpurple}{HTML}{F5F3FF}
\definecolor{lightgreen}{HTML}{F0FDF4}

\newcommand{\edgeicon}{\ensuremath{\heartsuit}}
\newcommand{\cloudicon}{\ensuremath{\diamondsuit}}
\newcommand{\trainicon}{\ensuremath{\clubsuit}}
\newcommand{\collabicon}{\ensuremath{\spadesuit}}

\newcommand{\meanstd}[2]{\ensuremath{#1_{\pm#2}}}

\newcommand{\method}{Sibyl\xspace}
\newcommand{\callcloud}{\texttt{call\_cloud}}

\title{
Sibyl: an efficient small--large model collaboration framework for long-horizon tasks
}

\author{
\bfseries
Zhewei Fang$^{1,*}$, Yuxin Zhang$^{2,*}$, Zhenwei Shao$^{1}$,
Mengze Li$^{3}$, Zheng Lin$^{4}$, Long Chen$^{5}$,\\
\bfseries
\space Zhou Yu$^{1,\dagger}$, Zhe Chen$^{2}$, Zhiwen Chen$^{6}$,
Zhaode Wang$^{6}$, Chengfei Lv$^{6}$\\[4pt]
\normalfont\small
$^{1}$Hangzhou Dianzi University \quad
$^{2}$Fudan University\\
$^{3}$The Hong Kong University of Science and Technology\\
$^{4}$University of Luxembourg \quad
$^{5}$Simon Fraser University\\
$^{6}$Alibaba Group\\[3pt]
\normalfont\small
$^{*}$Equal contribution.
\quad
$^{\dagger}$Corresponding author.
}

\iclrfinalcopy

\begin{document}

\maketitle
\fancyhead{}
\begin{abstract}
Small language models (SLMs) offer a promising foundation for on-device agents through low-latency, resource-efficient inference, yet limited reasoning and planning capabilities constrain their performance on long-horizon tasks requiring multi-step interaction with the environment.
Step-level collaboration between SLMs and larger cloud-hosted models can bridge this gap, but identifying states that warrant cloud assistance remains challenging: the contribution of each cloud call is entangled with subsequent actions and can be assessed only from the final task outcome. Compounding this challenge, the SLM must balance two competing objectives: maximizing task success and minimizing cloud calls.
To address this, we propose \method{}, an algorithm that trains SLM agents to selectively consult cloud models at the step level and internalize their guidance for subsequent decisions, achieving strong task performance with minimal cloud reliance.
\method{} follows a three-stage training pipeline that (1) builds a robust base policy through consultation-free self-evolving reinforcement learning (RL); (2) cold-starts consultation behavior via decisive-disagreement state mining; and (3) jointly optimizes consultation decisions and guidance internalization through consultation-aware RL.
Experiments on ALFWorld and WebShop demonstrate that \method{}, using only a 0.6B-parameter model, outperforms state-of-the-art baselines, including agent training and routing methods, by 95.2\% and 80.4\% in success rate while averaging only 0.8 and 3.9 cloud calls per trajectory, respectively.
We release the code and trained model weights at \url{https://github.com/real-fzw/Sibyl}
\end{abstract}

\section{Introduction}

Language model agents have recently shown strong promise on long-horizon tasks such as embodied interaction~\citep{seo2026from,kwon2026embodied}, web navigation~\citep{yang2025agentoccam,liu2026webcoach}, and software engineering~\citep{antoniades2025swesearch,zhang2025diversity}, which require sustained iteration over reasoning, tool use, observation, and revision across many interdependent steps~\citep{NEURIPS2023_1b44b878,xu2026theagentcompany}.
However, much of this progress has been driven by cloud-hosted large language models (LLMs) (e.g., GPT-5.6~\citep{gpt56}, Gemini 3.1 Pro~\citep{gemini31pro}, Qwen3.7-Max~\citep{qwen37}, and Kimi K3~\citep{kimik3}), while small language models (SLMs) suitable for edge deployment remain markedly less capable, particularly in the 3B-and-below regime~\citep{liu2024agentbench,mai2026agentic}.
Their limited reasoning and planning degrade decision quality over sustained interaction with the environment, undermining reliability on long-horizon tasks and posing a critical barrier to autonomous-agent deployment in latency- and resource-constrained applications that require efficient on-device execution~\citep{wan2025reca,wen2024autodroid}.

A natural solution is edge--cloud collaboration: on-device SLMs handle most steps locally and selectively consult stronger cloud-hosted LLMs~\citep{ong2025routellm,liaoalpharouter}, combining efficient local execution with advanced reasoning and planning.
However, collaboration on long-horizon agentic tasks is fundamentally more challenging than conventional SLM--LLM routing. First, model selection is no longer a one-time decision for each task, but a sequence of decisions throughout an evolving trajectory~\citep{zhang2026hera}. Consequently, the utility of step-level consultation is entangled with subsequent actions and often discernible only from the final outcome, complicating the precise identification of beneficial consultation states required to maximize task success with minimal cloud calls~\citep{feng2025gigpo}.
Second, existing SLM--LLM collaboration methods typically rely on an external router~\citep{ong2025routellm,chen2024frugalgpt}, which is ill-suited to long-horizon agents: routing decisions are tightly coupled with the agent's acting policy, making it difficult to train a separate router that generalizes across diverse, evolving states and remains aligned as the policy is updated. Moreover, a separate router adds supervision overhead and system complexity, complicating end-to-end agent training.

\begin{figure}[t]
    \centering
    \includegraphics[width=\linewidth]{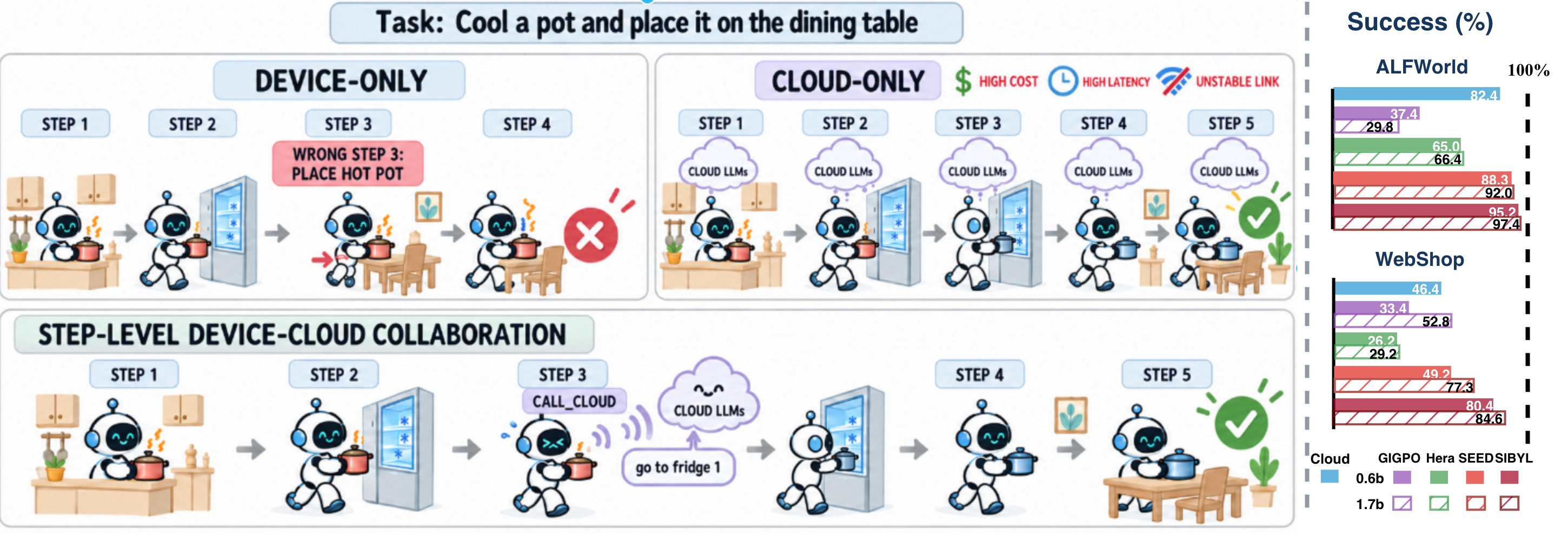}
    \caption{\method{} enables step-level device--cloud collaboration on long-horizon tasks, combining high success with low latency and cost. Right: success-rate comparisons with representative baselines; see \S~\ref{sec:eval} for further results.}
    \label{fig:teaser}
\end{figure}

We address these challenges with \method{}, a policy optimization method that trains SLM agents to selectively consult cloud-hosted LLMs and internalize their guidance; Figure~\ref{fig:teaser} illustrates the step-level collaboration paradigmd. Without a separate router, \method{} treats consultation as an SLM action and optimizes it alongside local actions from trajectory-level outcomes, while progressively reducing cloud reliance through guidance internalization.
\method{} training proceeds in three stages:
(1) consultation-free self-evolving reinforcement learning (RL) distills hindsight skills from on-policy trajectories to strengthen SLM agent before introducing cloud assistance;
(2) decisive-disagreement state mining identifies states where a one-step cloud intervention turns a failed SLM trajectory into a successful one, providing targeted supervision to cold-start consultation behavior;
(3) consultation-aware RL jointly optimizes local actions and cloud calls using trajectory-level outcomes while internalizing LLM guidance through hindsight skill distillation, thereby improving task success and reducing cloud reliance.

Experiments on challenging ALFWorld and WebShop benchmarks demonstrate that \method{} establishes a new task-success--execution-cost Pareto frontier relative to both state-of-the-art agent training and routing methods. Notably, with only a 0.6B-parameter SLM, \method{} achieves success rates of 95.2\% and 80.4\% while averaging only 0.8 and 3.9 cloud calls per trajectory, respectively.

Our contributions are threefold:
\begin{itemize}
    \item We formulate long-horizon SLM--LLM collaboration as SLM policy optimization over local actions and step-level cloud consultation, eliminating the need for an external router.
    \item We propose \method{}, a three-stage training algorithm that enables SLM agents to learn when to ask for cloud assistance and internalize the resulting guidance, thereby minimizing cloud calls while preserving task success.
    \item Across challenging agent benchmarks, \method{} advances the success--cost Pareto frontier over state-of-the-art agent training and routing methods.
\end{itemize}

\section{Related Work}

\paragraph{Reinforcement learning for language-model agents.}
Language-model agents show substantial promise for autonomous decision-making by coupling reasoning with sustained environment interaction~\citep{yao2023react,NEURIPS2023_1b44b878}.
RL has become a central paradigm for training agents through direct policy improvement from interaction outcomes.
Building on the widely adopted clipped actor--critic objective~\citep{schulman2017ppo}, GRPO~\citep{shao2024deepseekmath} replaces the learned critic with group-relative advantage estimation.
GiGPO~\citep{feng2025gigpo} further introduces hierarchical trajectory- and step-level grouping for long-horizon credit assignment.
RAGEN~\citep{wang2025ragen} studies stability in self-evolving agentic RL, while SEED~\citep{wu2026seed} uses on-policy distillation (OPD) to distill hindsight skills from on-policy trajectories into the acting policy.
However, these methods optimize individual agents rather than multi-agent collaboration, making them less effective for capacity-limited SLMs, particularly those with \(<1\)B parameters.

\paragraph{SLM--LLM collaboration.}
Leveraging assistance from stronger LLMs has emerged as a prominent approach to mitigating SLM capacity limitations while balancing execution efficiency and cost~\citep{zhang2025satellite}.
FrugalGPT~\citep{chen2024frugalgpt} and Hybrid LLM~\citep{ding2024hybrid} route requests based on predicted quality--cost trade-offs, while Eagle~\citep{Eagle2024} and RouteLLM~\citep{ong2025routellm} use historical similarity or preference-based predictions.
MinionS instead assigns subtask decomposition and verification to the cloud model~\citep{narayan2025cost}. However, these methods target independent requests rather than evolving long-horizon trajectories. Most related to our work, Hera~\citep{zhang2026hera} extends collaboration to the step level but relies on a separately trained router over fixed device and cloud policies, leaving SLM capability unchanged. In contrast, \method{} integrates consultation and guidance internalization into the SLM policy, eliminating the external router while improving the system’s capability ceiling and cost efficiency.

\section{Problem Formulation}
\label{sec:formulation}

Given a long-horizon task $x \sim p(\mathcal{X})$, an on-device SLM agent with policy $\pi_{\theta^{\mathrm{S}}}$, parameterized by $\theta^{\mathrm{S}}$, interacts with the environment over $T$ steps.
At step $t$, the agent observes a state $s_t \in \mathcal{S}$ and selects an action $a_t \in \mathcal{A}(s_t)$ according to $a_t \sim \pi_{\theta^{\mathrm{S}}}(\cdot \mid s_t, x)$,
where $\mathcal{A}(s_t)$ denotes the admissible action set at state $s_t$. After executing $a_t$, the environment transitions to the next state $s_{t+1}$. A completed episode induces a trajectory:
\begin{equation}
\tau=\{(s_t,a_t)\}_{t=1}^{T},
\end{equation}
with a trajectory-level return $R(\tau)\in\mathbb{R}$.
In many agentic environments, $R(\tau)$ is sparse and observed only upon trajectory termination.
The training objective is to optimize policy parameters $\theta^{\mathrm{S}}$ to maximize expected return:
\begin{equation}
\max_{\theta^{\mathrm{S}}} \; \mathbb{E}_{x \sim p(\mathcal{X}),\, \tau \sim \pi_{\theta^{\mathrm{S}}}(\cdot \mid x)}[R(\tau)].
\end{equation}

To compensate for the SLM's limited capabilities, we augment its action space to enable one-step consultation with a stronger cloud-hosted LLM:
\begin{equation}
    \widetilde{\mathcal{A}}(s_t)
    = \mathcal{A}(s_t) \cup \{\callcloud\}.
\label{eq:add_action}
\end{equation}
If the agent selects an action in $\mathcal{A}(s_t)$, it acts locally as usual; if it selects $\texttt{call\_cloud}$, a stronger cloud-based policy $\pi_{\theta^{\mathrm{L}}}$ receives the current state $s_t$ and the task instruction $x$, and returns one executable action $\tilde a_t \in \mathcal{A}(s_t)$ according to $\tilde a_t \sim \pi_{\theta^{\mathrm{L}}}(\cdot \mid s_t, x)$ for step $t$.
The environment executes $\tilde a_t$, after which the local policy $\pi_{\theta^{\mathrm{S}}}$ resumes action selection.

Under this formulation, $\pi_{\theta^{\mathrm{S}}}$ jointly learns which local actions to take and when to consult the cloud $\pi_{\theta^{\mathrm{L}}}$, balancing expected task return against cloud-invocation cost. Let $C(\tau)$ denote the cumulative cost of $\pi_{\theta^{\mathrm{L}}}$ along trajectory $\tau$, such as API charges. We optimize the following cost-regularized policy objective:
\begin{equation}
\max_{\theta^{\mathrm{S}}}\;
\mathbb{E}_{x \sim p(\mathcal{X}),\,
\tau \sim \pi_{\theta^{\mathrm{S}}}^{\mathrm{coll}}(\cdot \mid x)}
\left[R(\tau)-\lambda C(\tau)\right].
\label{eq:optimization_goal}
\end{equation}
where $\lambda \ge 0$ controls the trade-off between task return and cloud cost, and $\pi_{\theta^{\mathrm{S}}}^{\mathrm{coll}}$ denotes the SLM policy augmented with $\callcloud$ action, and $\pi_{\theta^{\mathrm{L}}}$ remains fixed.

\section{Method}

\subsection{Overview}

\begin{figure}
    \centering \includegraphics[width=\linewidth]{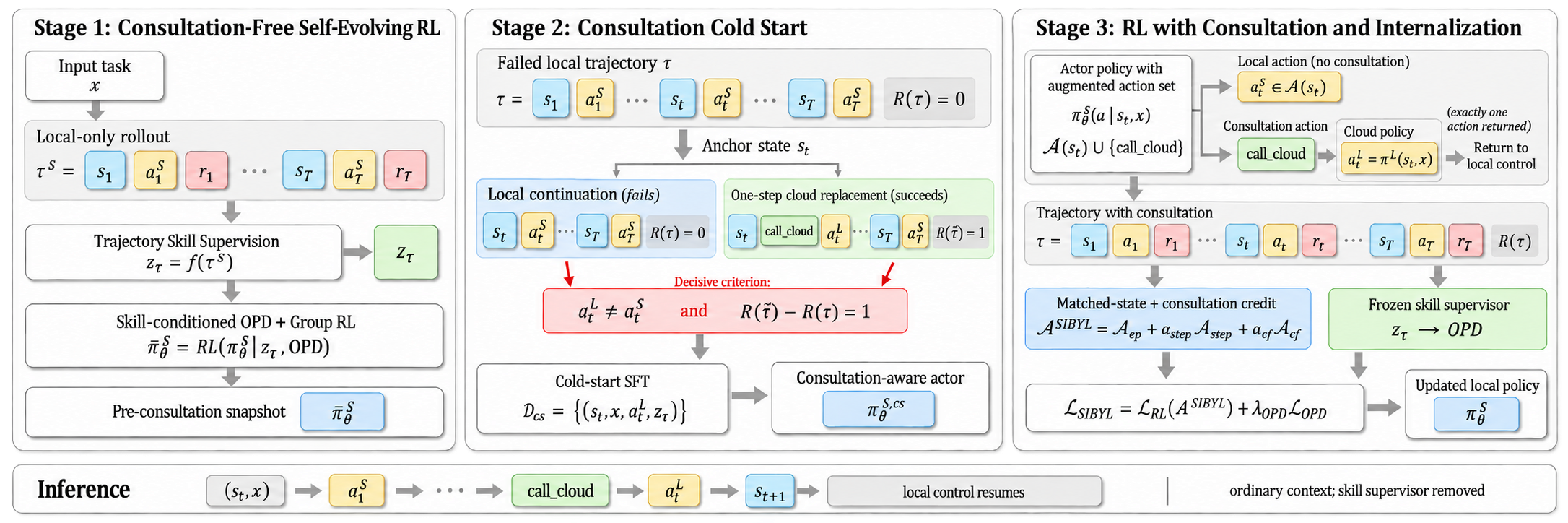}
    \caption{Three-stage training pipeline of \method{}, which trains an SLM to act locally, consult the cloud at critical steps, and internalize its guidance to achieve high task success with few cloud calls.}
    \label{fig:overview}
\end{figure}

Building on \S~\ref{sec:formulation}, we present \method{}, a method that trains SLM agents to selectively consult stronger cloud models and internalize their guidance to maximize task success while minimizing cloud calls.
\method{} follows a three-stage training pipeline, as illustrated in Figure~\ref{fig:overview}:
\begin{itemize}
    \item \textbf{Consultation-Free Self-Evolving RL (\S~\ref{sec:confree_rl}).} The SLM is first trained through RL with skill-conditioned on-policy distillation, providing a reliable basis for consultation.
    \item \textbf{Consultation Cold Start (\S~\ref{sec:concoldstart}).} Matched local and one-step cloud-assisted branches identify decisive-disagreement states to cold-start selective consultation.
    \item \textbf{RL with Consultation and Internalization (\S~\ref{sec:rl_co_in}).} On-policy RL uses step-level credit to jointly optimize local actions and consultation decisions while internalizing cloud guidance to strengthen local capability and reduce repeated calls.
\end{itemize}

\subsection{Consultation-Free Self-Evolving RL}
\label{sec:confree_rl}

Learning effective consultation requires a sufficiently capable base policy; otherwise, the SLM may improve task success by overusing cloud assistance rather than identifying states where consultation is truly necessary. We therefore first train $\pi_{\theta^{\mathrm{S}}}$ as a robust standalone agent without consultation. Inspired by self-evolving agent training~\citep{wu2026seed}, we distill reusable hindsight guidance from the agent's own trajectories, providing dense supervision beyond sparse outcome rewards for long-horizon behaviors such as subgoal planning and failure avoidance.

\paragraph{Trajectory-level skill supervision.}
We first execute training task $x$ using the SLM policy $\pi_{\theta^{\mathrm{S}}}$ alone, yielding a trajectory $\tau^{\mathrm{S}}$. Given $x$, $\tau^{\mathrm{S}}$, and its return $R(\tau^{\mathrm{S}})$, the cloud LLM $\pi_{\theta^{\mathrm{L}}}$ serves as a trajectory teacher $\mathcal{T}_{\theta^{\mathrm{L}}}$ to extract a concise causal summary and reusable hindsight skill:
\begin{equation}
z_{\tau^{\mathrm{S}}}=\mathcal{T}_{\theta^{\mathrm{L}}}(x,\tau^{\mathrm{S}},R(\tau^{\mathrm{S}})).
\end{equation}
Beyond the sparse return, $z_{\tau^{\mathrm{S}}}$ captures trajectory-level behavioral structure, such as effective subgoal ordering, decisive observations, and failure-avoidance rules. We then fine-tune $\pi_{\theta^{\mathrm{S}}}$ on the resulting trajectory--skill pairs, using the trained model both as the trajectory analyzer $\mathcal{T}_{\theta^{\mathrm{S}}}$ for subsequent on-policy distillation and as the initialization of $\pi_{\theta^{\mathrm{S}}}$ itself. This bootstrapping eliminates repeated cloud queries for trajectory annotation during online training.

\paragraph{Skill-conditioned on-policy distillation.}
Since the trajectory--skill fine-tuning above trains $\pi_{\theta^{\mathrm{S}}}$ for trajectory analysis rather than action generation, we first restore its acting capability through a supervised cold start. We collect action trajectories from the stronger $\pi_{\theta^{\mathrm{L}}}$ and fine-tune $\pi_{\theta^{\mathrm{S}}}$ on standard state--action pairs $(s_t,x)\rightarrow a_t$, providing both protocol-consistent action generation and an initial capability transfer before on-policy optimization.

Next, we perform self-evolving RL with skill-conditioned on-policy distillation.
Specifically, for each task \(x_i\), cold-started \(\pi_{\theta^{\mathrm{S}}}\) generates a trajectory $\tau_i=\{(s_{i,t},a_{i,t})\}_{t=1}^{T_i}$, which is then analyzed by the trajectory analyzer $\mathcal{T}_{\theta^{\mathrm{S}}}$ to produce a trajectory-level hindsight skill \(z_i=\mathcal{T}_{\theta^{\mathrm{S}}}(x_i,\tau_i,R(\tau_i))\).
Each action \(a_{i,t}\) is a token sequence \(a_{i,t} = (y_{i,t,1}, \dots, y_{i,t,L_{i,t}})\) of length \(L_{i,t}\).
The sampled action tokens are kept fixed and re-scored by $\pi_{\theta^{\mathrm{S}}}$ under the ordinary context $(s_{i,t},x_i)$ and its skill-augmented counterpart $H(s_{i,t},x_i,z_i)$, yielding the respective log-probabilities:
\begin{align}
    \ell^{\varnothing}_{i,t,k}
    &= \log \pi_{\theta^{\mathrm{S}}}
       (y_{i,t,k}\mid s_{i,t},x_i,y_{i,t,<k}),\\
    \ell^{z}_{i,t,k}
    &= \log \pi_{\theta^{\mathrm{S}}}
       (y_{i,t,k}\mid H(s_{i,t},x_i,z_i),y_{i,t,<k}),
\end{align}
The skill-induced probability lift forms a detached confidence weight:
\begin{equation}
    w_{i,t,k}
    = \sigma\!\left(
        \beta_{\mathrm{opd}}
        \operatorname{sg}[\ell^{z}_{i,t,k}-\ell^{\varnothing}_{i,t,k}]
      \right),
\end{equation}
and the OPD objective is:
\begin{equation}
    \mathcal{L}_{\mathrm{opd}}(\theta^{\mathrm{S}})
    = \mathbb{E}_{i,t,k}
      \left[w_{i,t,k}
      \left(\operatorname{sg}[\ell^{z}_{i,t,k}]
      -\ell^{\varnothing}_{i,t,k}\right)\right],
    \label{eq:opd}
\end{equation}
where $\beta_{\mathrm{opd}}$ controls the confidence weighting sharpness, and
$\operatorname{sg}$ denotes stop-gradient.
Gradients flow only through the ordinary branch, enabling $\pi_{\theta^{\mathrm{S}}}$ to learn from online skill-supported behaviors.

The final consultation-free policy is jointly optimized with the OPD loss above and a group-based RL objective. For $N$ trajectories sampled from the same task $x_i$, we define the group-relative advantage as
$A_{i,n}=R(\tau_{i,n})-\frac{1}{N}\sum_{n'=1}^{N}R(\tau_{i,n'})$.
Let $\ell^{\mathrm{old}}_{i,n,t,k}$ denote the token log-probability under the rollout policy and
$\rho_{i,n,t,k}(\theta^{\mathrm{S}})=\exp(\ell^{\varnothing}_{i,n,t,k}-\ell^{\mathrm{old}}_{i,n,t,k})$
the corresponding importance ratio. Following GRPO~\citep{shao2024deepseekmath}, the RL loss is:
\begin{equation}
\mathcal{L}_{\mathrm{rl}}(\theta^{\mathrm{S}})
=
-\mathbb{E}_{i,n,t,k}
\left[
\min\!\left(
\rho_{i,n,t,k}(\theta)A_{i,n},
\operatorname{clip}(\rho_{i,n,t,k}(\theta),1-\epsilon,1+\epsilon)A_{i,n}
\right)
\right]
+
\beta_{\mathrm{KL}} D_{\mathrm{KL}},
\end{equation}
where $\epsilon$ and $\beta_{\mathrm{KL}}$ control policy clipping and KL regularization, respectively.

The full objective of the consultation-free stage in \method{} is:
\begin{equation}
    \mathcal{L}_{\mathrm{confree}}
    =
    \mathcal{L}_{\mathrm{rl}}
    +
    \lambda_{\mathrm{opd}}\mathcal{L}_{\mathrm{opd}}.
\end{equation}
Here, $\mathcal{L}_{\mathrm{rl}}$ improves task-level performance from environment feedback, while $\mathcal{L}_{\mathrm{opd}}$ supplies dense hindsight supervision on the same on-policy trajectories, with $\lambda_{\mathrm{opd}}$ controlling the OPD signal strength.
After the complete self-evolving RL in this section, we freeze the resulting checkpoint as $\pi_{\bar\theta^{\mathrm{S}}}$, which serves as the pre-consultation snapshot for subsequent training with active consultation and consultation-feedback internalization.

\subsection{Consultation Cold Start}
\label{sec:concoldstart}
This section introduces \callcloud{} as a new action for $\pi_{\theta^{\mathrm{S}}}$ (Eq.~\ref{eq:add_action}) and provides an explicit cold start for its use.

\paragraph{Decisive-Disagreement State Mining.}

The pre-consultation policy $\pi_{\bar\theta^{\mathrm{S}}}$ is first rolled out on each training task $x_i$ to collect the failed-trajectory set:
\begin{equation}
\mathcal{D}^{-}
=
\{\tau_i \mid R(\tau_i)=0,\ \tau_i \sim \pi_{\bar\theta^{\mathrm{S}}}(\cdot \mid x_i)\},
\qquad
\tau_i=\{(s_{i,t},a^{\mathrm{S}}_{i,t})\}_{t=1}^{T_i}.
\end{equation}
For each $s_{i,t}$ visited in $\mathcal{D}^{-}$, the environment instance, action history, current observation, and admissible action set are recorded and fed to the cloud policy $\pi_{\theta^{\mathrm{L}}}$ to perform a one-step action:
\begin{equation}
a_{i,t}^{\mathrm{L}} \sim \pi_{\theta^{\mathrm{L}}}(\cdot \mid s_{i,t},x_i).
\end{equation}
If $a_{i,t}^{\mathrm{L}}=a_{i,t}^{\mathrm{S}}$, the cloud and local policies agree on this step; otherwise, a disagreement branch is formed by replacing $a_{i,t}^{\mathrm{S}}$ with $a_{i,t}^{\mathrm{L}}$ at state $s_{i,t}$, after which the subsequent steps are executed by $\pi_{\bar\theta^{\mathrm{S}}}$:
\begin{equation}
\widetilde{\tau}_{i}
=
\{(s_{i,1},a_{i,1}^{\mathrm{S}}),\dots,(s_{i,t-1},a_{i,t-1}^{\mathrm{S}}),(s_{i,t},a_{i,t}^{\mathrm{L}}),(s_{i,t+1},\widetilde a_{i,t+1}),\dots,(s_{i,\widetilde{T}_{i,t}},\widetilde a_{i,\widetilde{T}_{i,t}})\},
\end{equation}
where $\widetilde a_{i,u}\sim\pi_{\bar\theta^{\mathrm{S}}}(\cdot\mid s_{i,u},x_i)$ for $u>t$. We define the set of \textit{decisive-disagreement} states as:
\[
\mathcal{S}_{\mathrm{D}}
=
\left\{
s_{i,t}
\mid
a_{i,t}^{\mathrm{L}} \neq a_{i,t}^{\mathrm{S}},
\;
R(\widetilde{\tau}_{i}) - R(\tau_{i}) = 1
\right\},
\]
identifying critical decision points where a single cloud intervention can reverse the outcome, thereby providing precise supervision for when to ask for cloud assistance.

\paragraph{Consultation Cold-Start Fine-Tuning.}
The mined decisive-disagreement states are converted into supervised trajectories for initializing the consultation behavior of $\pi_{\theta^{\mathrm{S}}}$.
For each $s_{i,t} \in \mathcal{S}_{\mathrm{D}}$, the original action history
\(
\{(s_{i,1},a_{i,1}^{\mathrm{S}}),\dots,(s_{i,t-1},a_{i,t-1}^{\mathrm{S}})\}
\)
is retained, and the target action at step $t$ is replaced by \callcloud{}.
The resulting trajectory with consultation is:
\begin{equation}
\tau_{i}^{\mathrm{call}}
=
\{(s_{i,1},a_{i,1}^{\mathrm{S}}),\dots,(s_{i,t-1},a_{i,t-1}^{\mathrm{S}}),(s_{i,t},\callcloud)\}.
\end{equation}
These consultation trajectories are mixed with successful consultation-free trajectories sampled from $\pi_{\bar\theta^{\mathrm{S}}}$ to fine-tune $\pi_{\theta^{\mathrm{S}}}$, cold-starting selective cloud consultation at decisive states while preserving its standalone acting capability.

\subsection{RL with Consultation and Internalization}
\label{sec:rl_co_in}
After consultation cold start, we continue training $\pi_{\theta^{\mathrm{S}}}$ with on-policy RL to jointly learn when to consult and internalize cloud guidance, enabling it to handle similar states independently while retaining consultation for cases beyond its capability.

\paragraph{Consultation-aware credit assignment.}
We retain the group-based RL objective $\mathcal{L}_{\mathrm{rl}}$ from \S~\ref{sec:confree_rl} to optimize $\pi_{\theta^{\mathrm{S}}}$ over the augmented action space. However, its trajectory-level advantage cannot disentangle the contribution of a consultation from subsequent actions. We therefore introduce consultation-aware credit by comparing cloud-assisted and local continuations from matched states.

Specifically, we construct an exact-state key $g_{i,n,t}$ for each visited state from its current observation and admissible action set. For each state $s_{i,n,t}$, we associate it with the matched-state set indexed by its key, formed over all states in the $N$ trajectories sampled for task $x_i$:
\begin{equation}
\mathcal{I}_{g_{i,n,t}}
=
\left\{
s_{i,m,u}
\;\middle|\;
1\leq m\leq N,\;
1\leq u\leq T_{i,m},\;
g_{i,m,u}=g
\right\},
\qquad
g=g_{i,n,t}.
\end{equation}
Let $\mu_{i,n,t}$ denote the mean trajectory return of the states in $\mathcal{I}_{g_{i,n,t}}$. We define the matched-state advantage as
$A^{\mathrm{step}}_{i,n,t}=R(\tau_{i,n})-\mu_{i,n,t}$.

To further isolate the contribution of consultation, for
$q\in\{\mathrm{call},\mathrm{local}\}$, we define the action-conditioned mean return within $\mathcal{I}_g$ as:
\begin{equation}
\overline{R}^{q}_g
=
\frac{
\sum_{s_{i,m,u}\in\mathcal{I}_g}
\mathbb{I}_{q}(a_{i,m,u})R(\tau_{i,m})
}{
\sum_{s_{i,m,u}\in\mathcal{I}_g}
\mathbb{I}_{q}(a_{i,m,u})
},
\end{equation}
where $\mathbb{I}_{\mathrm{call}}(a)=1$ if $a=\callcloud$, and
$\mathbb{I}_{\mathrm{local}}(a)=1$ if $a\in\mathcal{A}(s)$. For each consultation state $s_{i,n,t}$, we define:
\begin{equation}
A^{\mathrm{cf}}_{i,n,t}
=
\overline{R}^{\mathrm{call}}_{g_{i,n,t}}
-
\overline{R}^{\mathrm{local}}_{g_{i,n,t}}.
\label{eq:consultadv}
\end{equation}
Notably, $A^{\mathrm{cf}}_{i,n,t}=0$ for local actions or when no matched local continuation exists. Combining this consultation-specific term with the trajectory- and matched-state-level advantages, we define
\begin{equation}
A^{\method}_{i,n,t}
=
A_{i,n}
+
\alpha_{\mathrm{step}}A^{\mathrm{step}}_{i,n,t}
+
\alpha_{\mathrm{cf}}A^{\mathrm{cf}}_{i,n,t},
\label{eq:totaladv}
\end{equation}
where $\alpha_{\mathrm{step}}$ and $\alpha_{\mathrm{cf}}$ control the contributions of matched-state and consultation-specific credit, respectively. This formulation preserves trajectory-level optimization while assigning finer-grained credit to matched states and beneficial consultations.

\paragraph{Cloud-guidance internalization.}
To further reduce cloud cost while preserving task success, we internalize guidance from on-policy cloud-assisted trajectories, enabling $\pi_{\theta^{\mathrm{S}}}$ to handle similar states independently whenever possible while retaining consultation for cases beyond its capability.

We reuse the frozen trajectory-skill supervisor $\mathcal{T}_{\theta^{\mathrm{S}}}$ from \S~\ref{sec:confree_rl}. After each rollout, $\mathcal{T}_{\theta^{\mathrm{S}}}$ analyzes the trajectory $\tau_{i,n}$, including consultation decisions, executed cloud actions, subsequent observations, and the final outcome, to extract a hindsight skill:
\begin{equation} z_{i,n} = \mathcal{T}_{\theta^{\mathrm{S}}}\bigl(x_i,\tau_{i,n},R(\tau_{i,n})\bigr). \end{equation}
The skill-conditioned $\mathcal{L}_{\mathrm{opd}}$ from \S~\ref{sec:confree_rl} (Eq.~\ref{eq:opd}) is then applied to distill behaviors favored under $z_{i,n}$ into the ordinary policy.

Since $z_{i,n}$ is extracted from on-policy trajectories, cloud guidance internalization tracks the evolving state distribution.

Combining the above components, the final-stage training objective of \method is: \begin{equation} \mathcal{L}_{\method} = \mathcal{L}_{\mathrm{rl}} \bigl(\theta^{\mathrm{S}};A^{\method}\bigr) + \lambda_{\mathrm{opd}}\mathcal{L}_{\mathrm{opd}}, \end{equation}
where $\lambda_{\mathrm{opd}}$ controls the strength of guidance internalization.
At inference time, the trajectory-skill supervisor and skill-augmented branch are removed.

Through the full three-stage training pipeline, \method enables SLM agents to seek cloud assistance when beneficial and internalize reusable guidance to strengthen local capability, thereby further reducing cloud cost while preserving task success.
Algorithm~\ref{alg:algorithm} details the training procedure.

Together, matched-state credit and guidance internalization indirectly pursue the success–cost objective in Eq.~\ref{eq:optimization_goal} by focusing consultations on consequential states and reducing repeated cloud use. Tuning $\alpha_{\mathrm{cf}}$ provides an implicit control over the performance–cost balance represented by $\lambda$ in Eq.~\ref{eq:optimization_goal}, producing the empirical Pareto frontier shown in \S~\ref{sec:eval} and Figure~\ref{fig:tradeoff}.

\section{Experiments}
\label{sec:eval}
This section empirically evaluate \method across a variety of long-horizon agentic tasks, aiming to assess task performance, consultation efficiency, the contributions of each training stage, and training and inference costs.

\begin{table*}[t]
\caption{
Results on ALFWorld and WebShop, averaged over three runs. \edgeicon, \(\cloudicon\), \(\trainicon\), and \(\collabicon\): device-only, cloud-only, cloud-assisted training, and collaborative inference, respectively.
The Qwen3-1.7B results for GRPO and SEED are from~\citep{wu2026seed}.
}
\centering
\scriptsize
\setlength{\tabcolsep}{0pt}
\renewcommand{\arraystretch}{1.02}

\begin{tabularx}{\textwidth}{
@{}
>{\raggedright\arraybackslash}p{0.24\textwidth}
*{7}{>{\centering\arraybackslash}X}
|
*{2}{>{\centering\arraybackslash}X}
@{}
}

\toprule
\multirow[c]{2}{*}[-0.6ex]{\textbf{Method}}
& \multicolumn{7}{c}{\textbf{ALFWorld}}
& \multicolumn{2}{c}{\textbf{WebShop}} \\
\cmidrule(lr){2-10}
& Pick & Look & Clean & Heat & Cool & Pick2 & All
& Score & Succ. \\
\midrule

\multicolumn{10}{l}{\textit{\textcolor{black!70}{Closed-Source Model}}} \\

\cloudicon\,Qwen3.7-Max
& \meanstd{88.6}{2.9}
& \meanstd{100.0}{0.0}
& \meanstd{80.3}{7.7}
& \meanstd{66.7}{3.6}
& \meanstd{64.0}{4.0}
& \meanstd{95.8}{4.2}
& \meanstd{82.4}{0.8}
& \meanstd{52.7}{0.2}
& \meanstd{46.4}{0.6}
\\

\cloudicon\,Gemini 3.1 Pro
& \meanstd{100.0}{0.0}
& \meanstd{100.0}{0.0}
& \meanstd{96.3}{3.7}
& \meanstd{100.0}{0.0}
& \meanstd{100.0}{0.0}
& \meanstd{95.8}{4.1}
& \meanstd{98.6}{1.4}
& \meanstd{65.2}{0.2}
& \meanstd{60.0}{0.4}
\\

\midrule

\multicolumn{10}{l}{
\textit{\textcolor{black!70}{
Qwen3-0.6B assisted by Qwen3.7-Max
}}
} \\

\edgeicon\,GRPO
& \meanstd{12.4}{1.7}
& \meanstd{25.6}{19.4}
& \meanstd{6.2}{5.7}
& \meanstd{4.2}{3.6}
& \meanstd{6.7}{2.3}
& \meanstd{4.2}{4.2}
& \meanstd{9.1}{2.3}
& \meanstd{67.1}{0.1}
& \meanstd{16.0}{0.4}
\\

\edgeicon\,GiGPO
& \meanstd{54.3}{2.9}
& \meanstd{59.0}{4.4}
& \meanstd{46.9}{7.7}
& \meanstd{27.1}{13.0}
& \meanstd{33.3}{14.1}
& \meanstd{1.4}{2.4}
& \meanstd{37.4}{3.2}
& \meanstd{78.6}{7.4}
& \meanstd{33.4}{6.8}
\\
\trainicon\,GRPO$_{w/ \text{SFT}}$
& \meanstd{100.0}{0.0}
& \meanstd{87.2}{4.4}
& \meanstd{100.0}{0.0}
& \meanstd{79.2}{3.6}
& \meanstd{78.7}{6.1}
& \meanstd{88.9}{2.4}
& \meanstd{90.7}{1.2}
& \meanstd{85.9}{1.1}
& \meanstd{79.8}{0.9}
\\

\trainicon\,GiGPO$_{w/ \text{SFT}}$
& \meanstd{100.0}{0.0}
& \meanstd{82.1}{11.7}
& \meanstd{98.8}{2.1}
& \meanstd{85.4}{13.0}
& \meanstd{82.7}{4.6}
& \meanstd{90.3}{6.4}
& \meanstd{91.7}{0.4}
& \meanstd{86.7}{1.3}
& \meanstd{80.6}{0.6}
\\

\trainicon\,SEED
& \meanstd{84.8}{4.3}
& \meanstd{89.7}{4.4}
& \meanstd{92.6}{6.4}
& \meanstd{93.8}{6.2}
& \meanstd{85.3}{12.9}
& \meanstd{87.5}{4.2}
& \meanstd{88.3}{0.4}
& \meanstd{67.6}{0.1}
& \meanstd{49.2}{1.4}
\\
\collabicon\,RouteLLM
& \meanstd{77.1}{2.9}
& \meanstd{84.6}{7.7}
& \meanstd{51.9}{6.4}
& \meanstd{62.5}{6.3}
& \meanstd{44.0}{4.0}
& \meanstd{79.2}{4.2}
& \meanstd{65.7}{1.9}
& \meanstd{24.6}{1.2}
& \meanstd{22.8}{0.8}
\\

\collabicon\,Hera
& \meanstd{85.7}{2.9}
& \meanstd{71.8}{4.4}
& \meanstd{33.3}{3.7}
& \meanstd{60.4}{7.2}
& \meanstd{65.3}{12.9}
& \meanstd{69.4}{2.4}
& \meanstd{65.0}{2.1}
& \meanstd{32.7}{1.7}
& \meanstd{26.2}{1.2}
\\

\rowcolor{lightgray}
\collabicon\,\textbf{\method}
& \meanstd{100.0}{0.0}
& \meanstd{92.3}{0.0}
& \meanstd{100.0}{0.0}
& \meanstd{89.6}{3.6}
& \meanstd{92.0}{0.0}
& \meanstd{91.7}{4.2}
& \meanstd{95.2}{0.8}
& \meanstd{90.7}{0.3}
& \meanstd{80.4}{0.5}
\\

\midrule

\multicolumn{10}{l}{
\textit{\textcolor{black!70}{
Qwen3-1.7B assisted by Qwen3.7-Max
}}
} \\

\edgeicon\,GRPO
& 71.1 & 41.7 & 36.4 & 40.0 & 31.8 & 31.6 & 46.1 & 67.3 & 38.3 \\

\edgeicon\,GIGPO
& \meanstd{48.6}{7.5}
& \meanstd{33.3}{4.4}
& \meanstd{35.8}{2.1}
& \meanstd{25.0}{6.2}
& \meanstd{21.3}{9.2}
& \meanstd{5.6}{2.4}
& \meanstd{29.8}{2.8}
& \meanstd{53.9}{0.1}
& \meanstd{52.8}{0.4}
\\
\trainicon\,GRPO$_{w/ \text{SFT}}$
& \meanstd{100.0}{0.0}
& \meanstd{87.2}{8.9}
& \meanstd{96.3}{3.7}
& \meanstd{89.6}{13.0}
& \meanstd{88.0}{14.4}
& \meanstd{52.8}{4.8}
& \meanstd{86.7}{4.8}
& \meanstd{90.0}{0.8}
& \meanstd{72.2}{1.2}
\\
\trainicon\,GiGPO$_{w/ \text{SFT}}$
& \meanstd{100.0}{0.0}
& \meanstd{94.9}{4.4}
& \meanstd{96.3}{3.7}
& \meanstd{97.9}{3.6}
& \meanstd{94.7}{4.6}
& \meanstd{61.1}{4.8}
& \meanstd{91.0}{1.1}
& \meanstd{89.0}{0.5}
& \meanstd{75.4}{0.4}
\\

\trainicon\,SEED
& 97.6 & 100.0 & 100.0 & 80.0 & 84.2 & 90.0 & 92.0 & 87.1 & 77.3 \\
\collabicon\,RouteLLM
& \meanstd{84.8}{1.7}
& \meanstd{87.2}{4.4}
& \meanstd{38.3}{2.1}
& \meanstd{39.6}{3.6}
& \meanstd{40.0}{6.9}
& \meanstd{63.9}{8.7}
& \meanstd{59.3}{3.6}
& \meanstd{27.2}{1.4}
& \meanstd{24.4}{3.6}
\\

\collabicon\,Hera
& \meanstd{92.4}{1.7}
& \meanstd{87.2}{8.9}
& \meanstd{27.2}{4.3}
& \meanstd{43.8}{16.5}
& \meanstd{72.0}{8.0}
& \meanstd{70.8}{8.3}
& \meanstd{66.4}{2.1}
& \meanstd{38.6}{1.8}
& \meanstd{29.2}{0.6}
\\

\rowcolor{lightgray}
\collabicon\,\textbf{\method}
& \meanstd{100.0}{0.0}
& \meanstd{100.0}{0.0}
& \meanstd{100.0}{0.0}
& \meanstd{100.0}{0.0}
& \meanstd{92.0}{4.0}
& \meanstd{93.1}{2.4}
& \meanstd{97.4}{0.8}
& \meanstd{91.1}{0.9}
& \meanstd{84.6}{0.6}
\\

\bottomrule
\end{tabularx}
\label{tab:main_results}
\end{table*}

\subsection{Experimental Setup}
\paragraph{Benchmarks.}
We evaluate \method{} on challenging benchmarks including ALFWorld~\citep{shridhar2021alfworld} and WebShop~\citep{yao2022webshop}.
ALFWorld is a text-based embodied benchmark with 3,827 instances across six household task types: Pick \& Place (Pick), Examine in Light (Look), Clean \& Place (Clean), Heat \& Place (Heat), Cool \& Place (Cool), and Pick Two \& Place (Pick2). WebShop requires agents to navigate a simulated e-commerce site and purchase suitable products from over 1.1 million items based on 12K user instructions. Our detailed dataset splits are provided in Appendix~\ref{app:data}.

\paragraph{Training details.}
We train Qwen3-0.6B/1.7B~\citep{yang2025qwen3} as the on-device SLMs, assisted by Qwen3.7-Max~\citep{qwen37} and Gemini 3.1 Pro~\citep{gemini31pro} as cloud LLMs.
Agent outputs follow the ReAct~\citep{yao2023react} format.
Training is conducted on eight AMD Instinct MI308X GPUs.
Stage 1 (\S~\ref{sec:confree_rl}) trains $\mathcal{T}_{\theta^{\mathrm{S}}}$ with learning rate (lr) $5\times10^{-6}$ and batch size (bs) 8,
then performs consultation-free RL with $\pi_{\theta^{\mathrm{S}}}$ at lr $10^{-6}$ with group size 8, bs 256, $\beta_{\mathrm{opd}}=5.0$, and $\lambda_{\mathrm{opd}}=0.01$.
Stage 2 (\S~\ref{sec:concoldstart}) trains at lr $2\times10^{-6}$ with bs 16.
Stage 3 (\S~\ref{sec:rl_co_in}) performs RL at lr $10^{-6}$ with group size 8, bs 256, $\lambda_{\mathrm{opd}}=0.01$, $\alpha_{\mathrm{step}}=0.5$, and $\alpha_{\mathrm{cf}}=0.25$.
Further details are provided in Appendix~\ref{app:training}.

\paragraph{Baselines.}
We compare \method against four baseline families:
(1) \textit{cloud-only methods} directly prompt the frontier closed-source Qwen3.7-Max and Gemini 3.1 Pro;
(2) \textit{device-only methods} train SLMs with agentic RL, including GRPO~\citep{shao2024deepseekmath} with group-relative critic-free optimization and GiGPO~\citep{feng2025gigpo} with hierarchical step-level credit assignment;
(3) \textit{training-time cloud-assisted methods} include SEED~\citep{wu2026seed} with cloud-guided on-policy hindsight-skill distillation, and GRPO/GiGPO cold-started through supervised fine-tuning on cloud-generated trajectories;
(4) \textit{inference-time collaborative methods} include RouteLLM~\citep{ong2025routellm} and Hera~\citep{zhang2026hera}, which train routers for request-level and step-level routing, respectively.

\subsection{Overall Results}

Table~\ref{tab:main_results} shows that \method{} improves long-horizon task success across both SLM sizes. With Qwen3-0.6B, it achieves 95.2\% success on ALFWorld and 80.4\% on WebShop, versus 37.4\% and 33.4\% for the strongest device-only baselines. Cloud-assisted training substantially improves the SLM, but \method{} further raises ALFWorld success from 91.7\% to 95.2\% and achieves a higher WebShop score (90.7 versus 86.7); its WebShop success is comparable to the best training-time baseline (80.4\% versus 80.6\%). The higher score despite similar exact success suggests better partial progress on episodes that do not result in a complete purchase. With Qwen3-1.7B, \method{} exceeds the strongest training-time baselines by 5.4 and 7.3 success points on the two benchmarks, indicating that selective consultation remains useful as local model capacity increases. Among methods that use cloud assistance at inference time, \method{} achieves higher success than RouteLLM and Hera with fewer cloud calls, as detailed in the next section. Both baselines train an external router to allocate decisions between fixed device and cloud policies. \method{} instead jointly optimizes local actions and consultation within the SLM, enabling it to seek help at consequential states and internalize cloud guidance for similar future states. Notably, \method{} surpasses cloud-only Qwen3.7-Max on both benchmarks while consulting that same model for only a subset of steps, suggesting that it learns to combine local execution and targeted cloud intervention into more effective trajectories than cloud-only execution.

\begin{table*}[t]
\caption{
Task performance and test-time cloud cost on ALFWorld and WebShop. Call ratio is the percentage of executed steps delegated to the cloud, Calls/Ep. is the mean number of cloud calls per episode, and Cost is the total API cost over the full test set, calculated from cloud input/output token usage and API prices.
}
\centering
\scriptsize
\setlength{\tabcolsep}{3.0pt}
\renewcommand{\arraystretch}{1.02}
\resizebox{\textwidth}{!}{%
\begin{tabular}{lcccc|ccccc}
\toprule
\multirow[c]{2}{*}[-0.6ex]{\textbf{Method}}
& \multicolumn{4}{c|}{\textbf{ALFWorld}} & \multicolumn{5}{c}{\textbf{WebShop}} \\
\cmidrule(lr){2-10}
& Succ.$\uparrow$ & Call ratio$\downarrow$ & Calls / Ep.$\downarrow$ & Cost$\downarrow$ & Score$\uparrow$ & Succ.$\uparrow$ & Call ratio$\downarrow$ & Calls / Ep.$\downarrow$ & Cost$\downarrow$ \\
\midrule

\multicolumn{10}{l}{\textit{Closed-Source Model}} \\
Qwen3.7-Max
& 82.4 & 100.0 & 15.2 & \$8.6 & 52.7 & 46.4 & 100.0 & 10.8 & \$48.5 \\
Gemini-3.1 Pro
& 98.6 & 100.0 & 9.1 & \$5.8 & 65.2 & 60.0 & 100.0 & 8.4 & \$58.6 \\
\midrule

\multicolumn{10}{l}{\textit{Qwen3-0.6B assisted by Qwen3.7-Max}} \\
RouteLLM
& 65.7 & 64.5 & 12.5 & \$5.3 & 24.6 & 22.8 & 85.5 & 5.1 & \$18.3 \\

Hera
& 65.0 & 54.6 & 5.3 & \$4.8 & 32.7 & 26.2 & 68.9  & 4.2 & \$12.2 \\

\rowcolor{lightgray}

\textbf{\method}
& 95.2 & 7.7 & 0.8 & \$0.4
& 90.7 & 80.4 & 64.8 & 3.9 & \$16.1
\\

\midrule
\multicolumn{10}{l}{\textit{Qwen3-1.7B assisted by Qwen3.7-Max}} \\
RouteLLM
& 59.3 & 60.4 & 12.4 & \$5.1 & 27.2 & 24.4 & 86.6 & 5.7 & \$20.9 \\
Hera
& 66.4 & 47.9 & 4.8 & \$3.9 & 38.6 & 29.2 & 67.7  & 4.2 & \$12.0 \\

\rowcolor{lightgray}

\textbf{\method}
& 97.4 & 2.8 & 0.3 & \$0.2
& 91.1 & 84.6 & 3.1 & 0.2 & \$1.1 \\
\bottomrule 
\end{tabular}%
}
\label{tab:cloud_assist}
\end{table*}

\subsection{Analysis of Cloud Consultation}
\label{sec:cloud-consultation-analysis}

Table~\ref{tab:cloud_assist} shows that \method{} achieves strong task success with exceptionally low cloud-invocation cost. With a 0.6B SLM, it reaches 95.2\% success on ALFWorld and 80.4\% on WebShop while reducing cost by 21.5$\times$ and 3.0$\times$ relative to cloud-only Qwen3.7-Max. With a 1.7B SLM, success rises to 97.4\% and 84.6\%, at 43$\times$ and 44$\times$ lower cost; only 2.8\% and 3.1\% of steps invoke the cloud. The ALFWorld result approaches cloud-only Gemini 3.1 Pro (98.6\%), while the WebShop result substantially exceeds it (60.0\%). Although RouteLLM and Hera use external routers to coordinate fixed device and cloud policies, \method{} jointly learns local actions and consultation while internalizing cloud guidance. The SLM can thus handle similar states independently and reserve calls for cases beyond its capability, achieving a stronger success--cost trade-off.

\begin{figure}[t]
    \centering
    \begin{minipage}[t]{0.462\linewidth}
        \vspace{0pt}
        \centering

        \begin{minipage}[t]{0.49\linewidth}
            \centering
            \includegraphics[width=\linewidth]{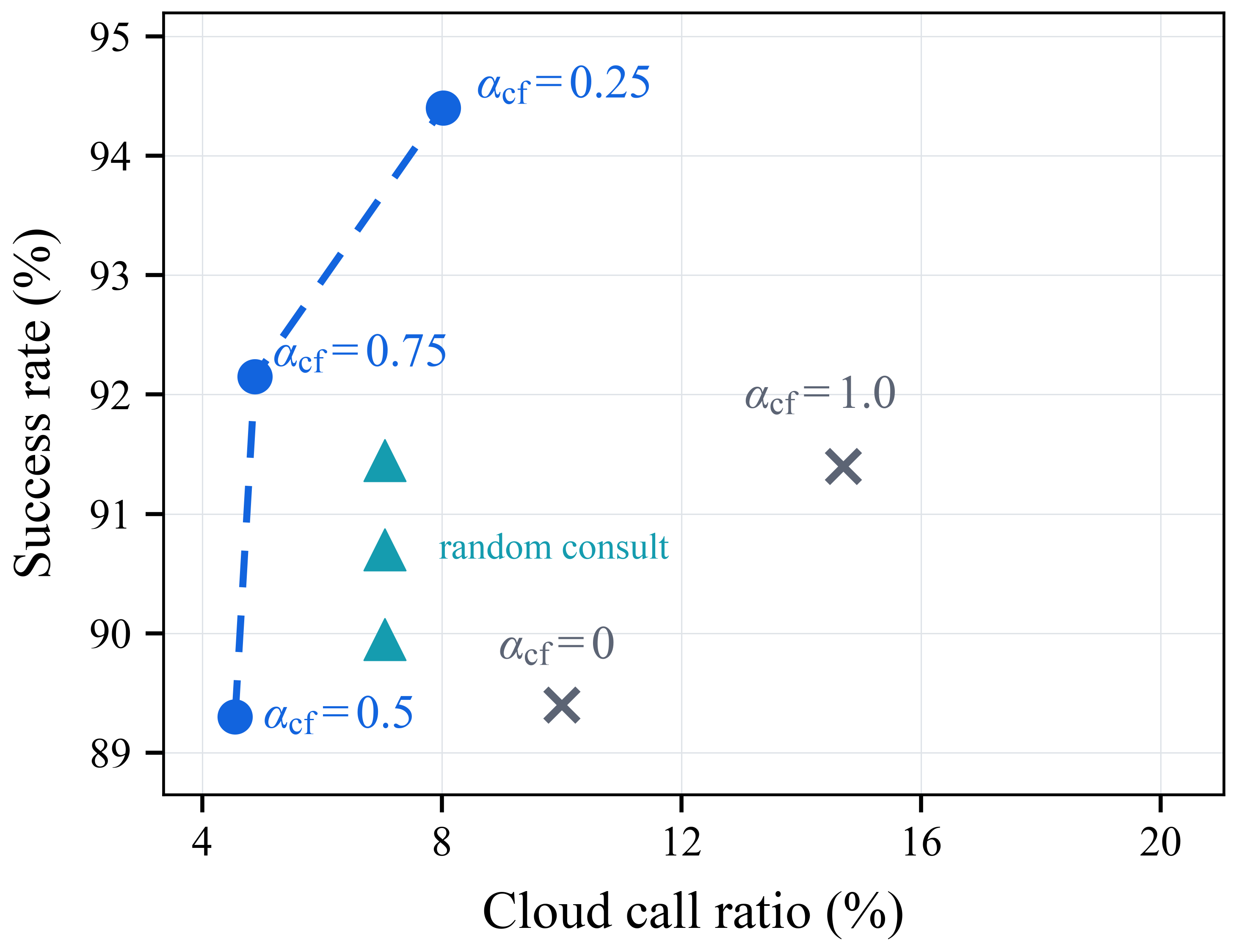}
            \par\smallskip
            {\small (a) ALFWorld}
        \end{minipage}
        \hfill
        \begin{minipage}[t]{0.49\linewidth}
            \centering
            \includegraphics[width=\linewidth]{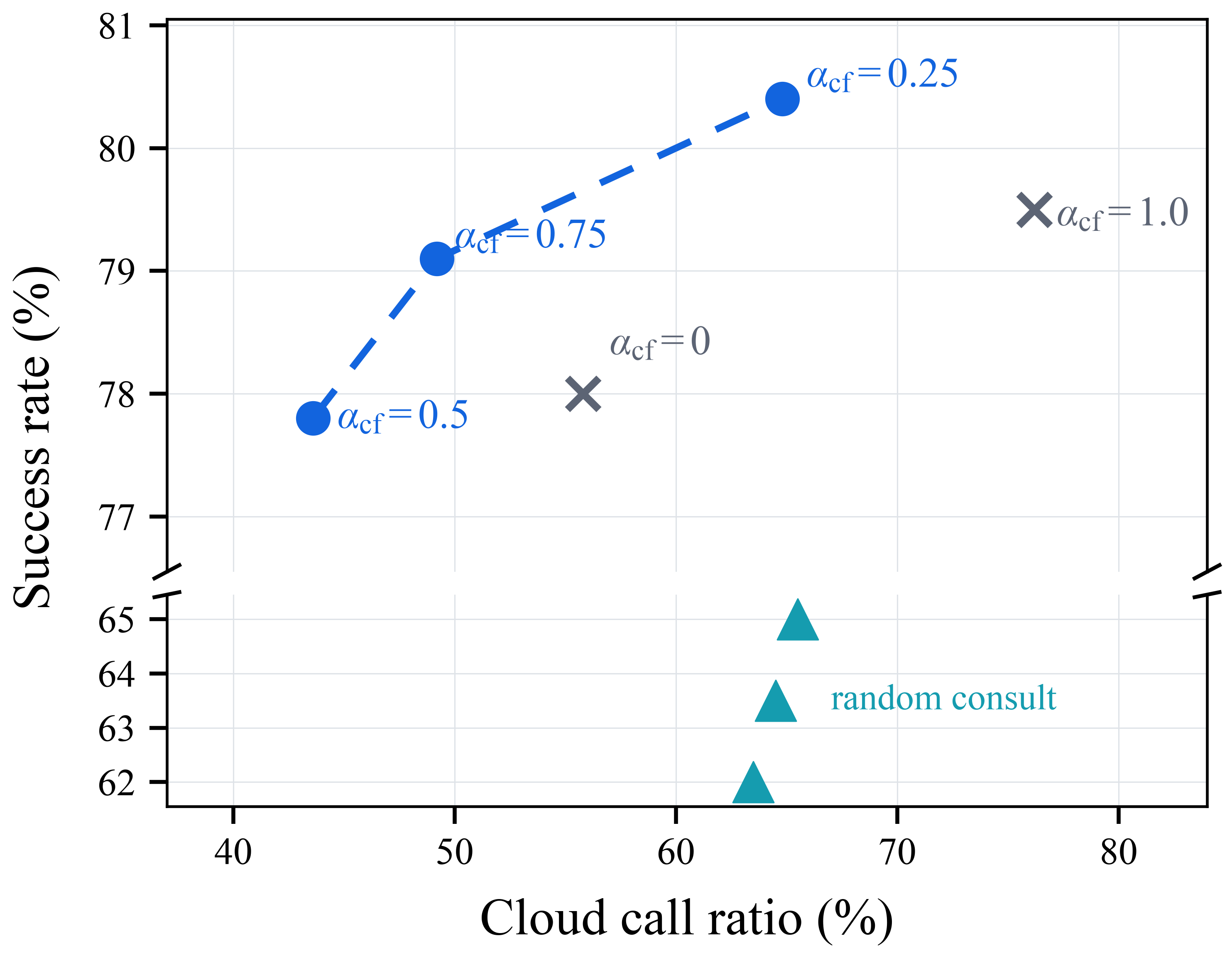}
            \par\smallskip
            {\small (b) WebShop}
        \end{minipage}

        \captionof{figure}{Success rate versus cloud call ratio on (a) ALFWorld and (b) WebShop.}
        \label{fig:tradeoff}
    \end{minipage}
    \hfill
    \begin{minipage}[t]{0.529\linewidth}
        \vspace{0pt}
        \centering

        \begin{minipage}[t]{0.49\linewidth}
            \centering
            \includegraphics[width=\linewidth]{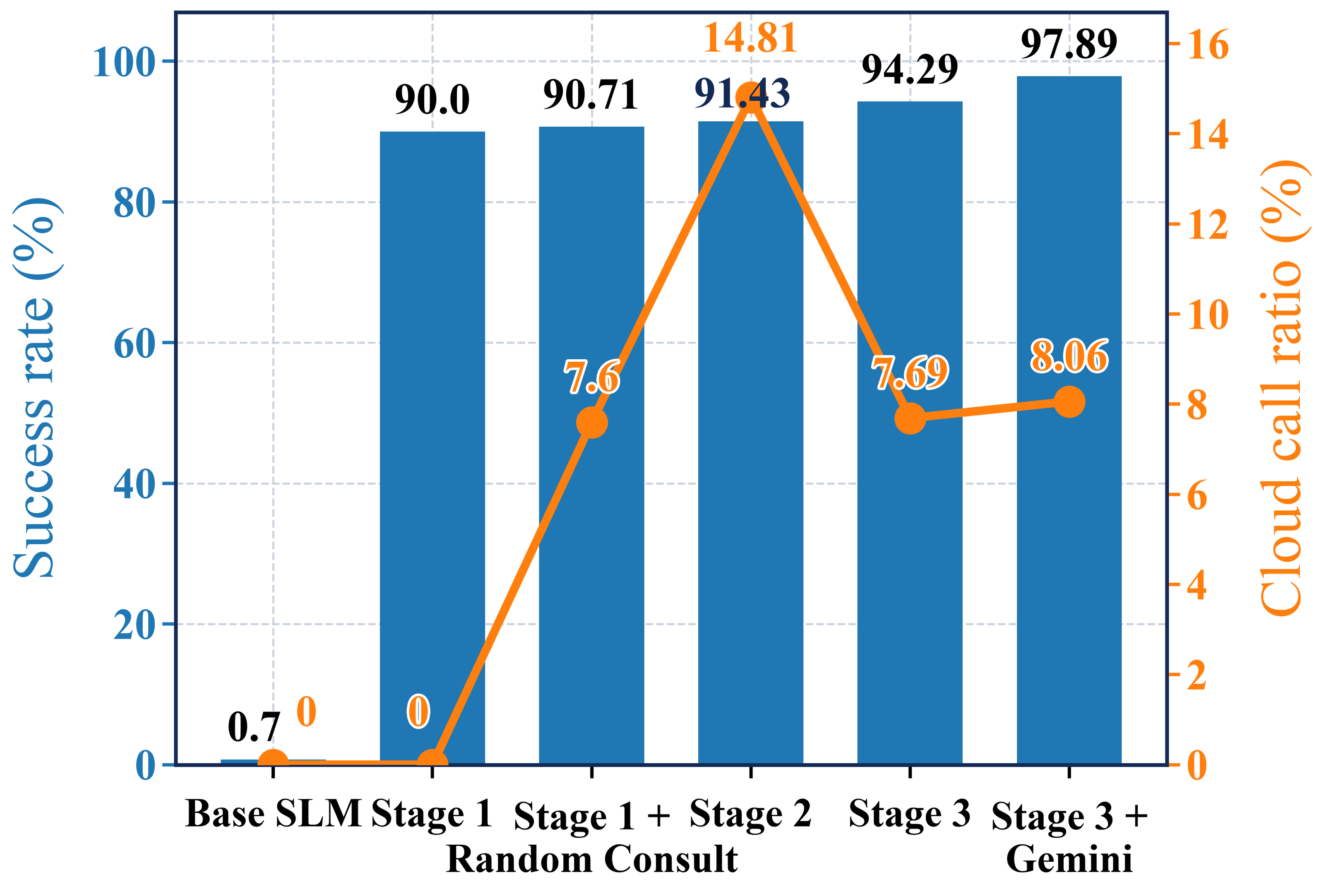}
            \par\smallskip
            {\small (a) ALFWorld}
        \end{minipage}
        \hfill
        \begin{minipage}[t]{0.49\linewidth}
            \centering
            \includegraphics[width=\linewidth]{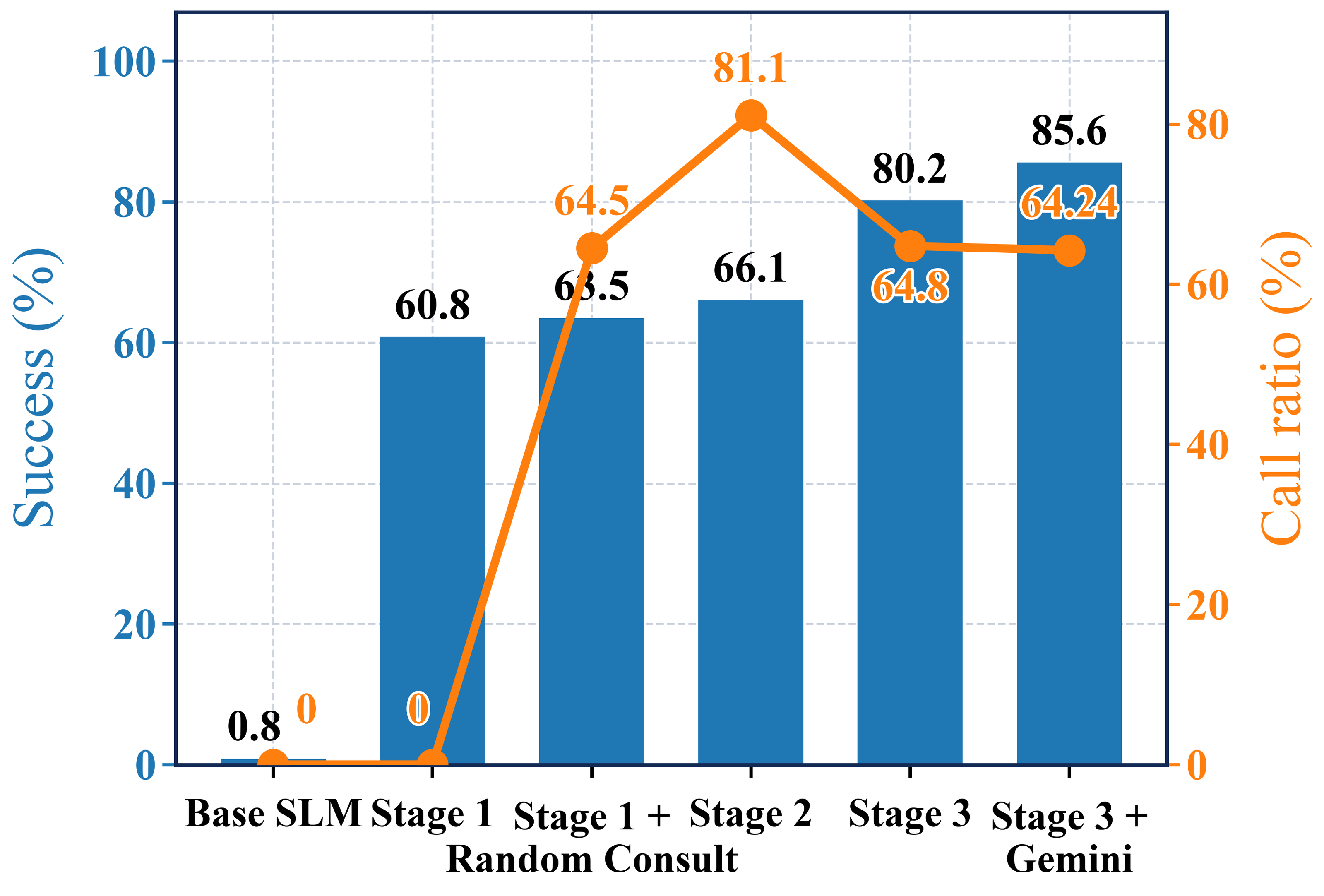}
            \par\smallskip
            {\small (b) WebShop}
        \end{minipage}

        \captionof{figure}{Ablation of \method{} with a Qwen3-0.6B SLM assisted by Qwen3.7-Max on (a) ALFWorld and (b) WebShop.}
        \label{fig:ablation}
    \end{minipage}
\end{figure}

Figure~\ref{fig:tradeoff} further shows that this advantage persists across cloud budgets. These results are from a first run  checkpoint, rather than the averages reported earlier. On ALFWorld, \method{} reaches 94.3\% success at an approximately 8\% call ratio, or 92.1\% with fewer than 5\% of steps delegated; the corresponding WebShop frontier reaches 80.4\% at 67\% and 77.8\% at 45\%. These settings outperform random consultation, RouteLLM, and Hera, indicating that the gains stem from learning when to seek assistance rather than simply having access to the cloud. Moreover, neither removing consultation-specific credit ($\alpha_{\mathrm{cf}}=0$) nor assigning it excessive weight ($\alpha_{\mathrm{cf}}=1$) reaches the frontier. Intermediate weights most effectively reward consequential consultations while suppressing unnecessary cloud calls.

\subsection{Ablation Study}
\label{sec:ablation}

Figure~\ref{fig:ablation} traces the contribution of each training stage on ALFWorld and WebShop. Starting from base SLM success rates of 0.7\% and 0.8\%, Stage 1 raises success to 90.0\% and 60.8\% without inference-time consultation, establishing a capable local policy. Adding random cloud calls yields only modest further gains, showing that cloud access alone is insufficient. Stage 2 improves success by cold-starting consultation at states where a one-step cloud action can reverse failure, but its higher call ratios indicate that this supervision has yet to learn when consultation can be avoided. Stage 3 effectively resolves this trade-off: success rises to 94.3\% and 80.2\% while call ratios fall to 7.7\% and 64.8\%, respectively. It also substantially outperforms random consultation at nearly the same call ratios, supporting the value of jointly optimizing consultation decisions and internalizing cloud guidance. Notably, replacing Qwen3.7-Max with the stronger Gemini 3.1 Pro only at inference improves success on both benchmarks without retraining the SLM, suggesting that the learned consultation policy can transfer across cloud models and benefit from stronger assistance.

\subsection{Cost Analysis}

\begin{wraptable}[7]{l}{0.49\textwidth}
    \vspace{-\intextsep}
    \centering
    \caption{Training and inference costs of \method{} on ALFWorld and WebShop.}
    \label{tab:beijing-api-cost}
    \scriptsize
    \setlength{\tabcolsep}{2.5pt}
    \renewcommand{\arraystretch}{1.15}

    \resizebox{\linewidth}{!}{%
    \begin{tabular}{lccc}
        \toprule
        & \shortstack{Training API Cost\\(Stage 1/2/3)}
        & \shortstack{Training\\GPU Hours}
        & \shortstack{Inference API Cost\\(full test set)} \\
        \midrule
        ALFWorld & \$10.7/\$1.7/\$92.7 & 149.7 & \$0.4 \\
        WebShop  & \$13.0/\$6.5/\$126.0 & 205.7 & \$16.1 \\
        \bottomrule
    \end{tabular}%
    }
\end{wraptable}

Table~\ref{tab:beijing-api-cost} reports training and inference API costs, together with training GPU-hours. Training incurs \$105.1 and \$145.5 in API charges for ALFWorld and WebShop, respectively, with Stage 3 accounting for the largest share. At inference time, \method reduces the average cloud API cost from \$8.6 to \$0.4 on ALFWorld and from \$48.5 to \$16.1 on WebShop, relative to cloud-only Qwen3.7-Max, corresponding to savings of 95.4\% and 66.8\%. The training API expenditure is therefore offset after approximately 1,794 ALFWorld or 2,245 WebShop tasks. These break-even estimates exclude the cost of 149.7 and 205.7 AMD Instinct MI308X GPU-hours of training, respectively.

\section{Conclusion}
We presented \method{} as a policy-optimization framework that trains SLM agents to selectively consult stronger cloud models and internalize their guidance.
By treating consultation as a native policy action, its three-stage training jointly strengthens local capability, learns when to ask, and reduces repeated cloud dependence.
Experiments on challenging long-horizon benchmarks show that \method{} achieves strong task success while substantially reducing cloud calls, advancing the success--cost trade-off over existing agent training and routing methods.
More broadly, \method{} offers an integrated framework that unifies coordination and capability transfer across heterogeneous language-model agents; its limitations and future directions are discussed in Appendix~\ref{app:limitations}.





\bibliography{references}
\bibliographystyle{iclr2027_conference}

\appendix

\newpage

\section{Pseudocode}
\label{alg:algorithm}

\begin{algorithm}[h]
\caption{\method{}: Three-Stage Training for Step-Level Device--Cloud Collaboration}
\label{alg:method}
\begin{algorithmic}[1]
\STATE \textbf{Input:} Training tasks $\mathcal{X}$; SLM policy $\pi_{\theta^{\mathrm{S}}}$; fixed cloud policy $\pi_{\theta^{\mathrm{L}}}$; rollouts per task $N$; weights $\lambda_{\mathrm{opd}},\alpha_{\mathrm{step}},\alpha_{\mathrm{cf}}$.
\STATE \textbf{Output:} Trained SLM policy $\pi_{\theta^{\mathrm{S}}}$.

\STATE \textbf{Stage 1: Consultation-free self-evolving RL}
\STATE Roll out $\pi_{\theta^{\mathrm{S}}}$ without consultation and obtain hindsight skills from the cloud trajectory teacher.
\STATE Fine-tune $\pi_{\theta^{\mathrm{S}}}$ on trajectory--skill pairs to initialize the skill supervisor $\mathcal{T}_{\theta^{\mathrm{S}}}$; restore action generation through SFT on cloud-generated state--action pairs.
\FOR{each RL iteration}
    \FORALL{tasks $x_i$}
        \STATE Sample $N$ local-only trajectories $\{\tau_{i,n}\}_{n=1}^{N}$.
        \STATE Extract $z_{i,n}=\mathcal{T}_{\theta^{\mathrm{S}}}(x_i,\tau_{i,n},R(\tau_{i,n}))$ and compute group-relative advantages $A_{i,n}$.
    \ENDFOR
    \STATE Update $\theta^{\mathrm{S}}$ using $\mathcal{L}_{\mathrm{rl}}+\lambda_{\mathrm{opd}}\mathcal{L}_{\mathrm{opd}}$.
\ENDFOR
\STATE Freeze the resulting policy as $\bar{\pi}^{\mathrm{S}}$ and the skill supervisor $\mathcal{T}_{\theta^{\mathrm{S}}}$.

\STATE \textbf{Stage 2: Consultation cold start}
\STATE Collect failed trajectories $\mathcal{D}^{-}$ and successful trajectories $\mathcal{D}^{+}$ using $\bar{\pi}^{\mathrm{S}}$.
\FORALL{states $s_{i,t}$ in trajectories $\tau_i\in\mathcal{D}^{-}$}
    \STATE Sample one cloud action $a^{\mathrm{L}}_{i,t}\sim\pi_{\theta^{\mathrm{L}}}(\cdot\mid s_{i,t},x_i)$.
    \IF{$a^{\mathrm{L}}_{i,t}\neq a^{\mathrm{S}}_{i,t}$}
        \STATE Replace $a^{\mathrm{S}}_{i,t}$ with $a^{\mathrm{L}}_{i,t}$ and complete the branch using $\bar{\pi}^{\mathrm{S}}$, obtaining $\tilde{\tau}_i$.
        \IF{$R(\tilde{\tau}_i)-R(\tau_i)=1$}
            \STATE Add the original trajectory prefix followed by $(s_{i,t},\texttt{call\_cloud})$ to $\mathcal{D}^{\mathrm{call}}$.
        \ENDIF
    \ENDIF
\ENDFOR
\STATE Fine-tune $\pi_{\theta^{\mathrm{S}}}$ on $\mathcal{D}^{\mathrm{call}}\cup\mathcal{D}^{+}$.

\STATE \textbf{Stage 3: RL with consultation and internalization}
\FOR{each RL iteration}
    \FORALL{tasks $x_i$}
        \STATE Sample $N$ trajectories using the augmented action space; each \texttt{call\_cloud} executes one cloud action before local control resumes.
        \STATE Group matching states by exact-state key $g$ and compute trajectory advantages $A_{i,n}$ and matched-state advantages $A^{\mathrm{step}}_{i,n,t}$.
        \STATE For consultation actions, compute $A^{\mathrm{cf}}_{i,n,t}=\overline{R}^{\mathrm{call}}_g-\overline{R}^{\mathrm{local}}_g$ when both groups exist; otherwise set it to zero.
        \STATE Set $A^{\mathrm{SIBYL}}_{i,n,t}=A_{i,n}+\alpha_{\mathrm{step}}A^{\mathrm{step}}_{i,n,t}+\alpha_{\mathrm{cf}}A^{\mathrm{cf}}_{i,n,t}$.
        \STATE Extract hindsight skills $z_{i,n}$ using the frozen $\mathcal{T}_{\theta^{\mathrm{S}}}$.
    \ENDFOR
    \STATE Update $\theta^{\mathrm{S}}$ using $\mathcal{L}_{\mathrm{rl}}(A^{\mathrm{SIBYL}})+\lambda_{\mathrm{opd}}\mathcal{L}_{\mathrm{opd}}$.
\ENDFOR
\RETURN $\pi_{\theta^{\mathrm{S}}}$.
\end{algorithmic}
\end{algorithm}

\section{Benchmark Protocol and Data Splits}
\label{app:data}
We evaluate the \method{} on two interactive benchmarks, ALFWorld and WebShop,
under the benchmark-specific protocols described below.

\paragraph{ALFWorld.}
ALFWorld~\citep{shridhar2021alfworld} provides a text-based view of household
tasks aligned with the embodied ALFRED environment. An agent receives a
natural-language goal, observes the environment through text, and selects
admissible actions until the goal is completed. The benchmark covers six task
types: \textit{Pick \& Place}, \textit{Examine in Light}, \textit{Clean \&
Place}, \textit{Heat \& Place}, \textit{Cool \& Place}, and \textit{Pick Two
\& Place}. The official split contains 3,553 training games, 140 Seen games,
and 134 Unseen games. Seen games use rooms encountered during training with
new object configurations, whereas Unseen games use previously unseen rooms
and layouts. We report the main results on the 140-task Seen split and use the
134-task Unseen split to assess out-of-distribution generalization. Each
episode is limited to 30 steps, and the prompt contains the most recent 5
interaction steps. The prompt and response limits are 2,048 and 512 tokens,
respectively.

\paragraph{WebShop.}
WebShop~\citep{yao2022webshop} is a simulated shopping website built from more
than 1.1 million products and 12,087 crowdsourced instructions. Each
instruction specifies requirements such as product attributes, options, and a
price limit. The agent searches the catalog, navigates product and detail
pages, selects options, and buys a product. The environment computes a reward
from the match between the selected product and the instruction, including its
type, attributes, options, and price. We report Task Score, defined as 100
times the mean reward, and Success Rate, the fraction of episodes with reward
one. We evaluate all methods on 500 test samples, with a maximum of 15 steps
per episode and the most recent 5 interaction steps included in the prompt.
The prompt and response limits are 12,288 and 512 tokens, respectively.

\section{Experimental Details}
\label{app:training}

\subsection{Prompts}
\label{app:prompts}

At each step, the on-device SLM receives the task instruction, current
observation, recent interaction history, and exact admissible actions. The
initial prompt contains no interaction history. The action set also includes
\texttt{call\_cloud}. For local action generation, both benchmarks use the
ReAct format with one \texttt{<think>} block followed by one
\texttt{<action>} block. ALFWorld actions are drawn from the admissible
environment actions, whereas WebShop actions take the form
\texttt{search[\ldots]} or \texttt{click[\ldots]}. When the SLM selects
\texttt{call\_cloud}, the cloud model receives the complete
action--observation--result trajectory and the current available actions, then
returns one executable action under the benchmark-specific prompts below.

\par\medskip
\noindent\textbf{ALFWorld.}\par
\smallskip

\begin{lstlisting}
You are an expert agent operating in the ALFRED Embodied Environment. Your task is to: {task_description}
Prior to this step, you have already taken {step_count} step(s). Below are the most recent {history_length} observations and the corresponding actions you took: {action_history}
You are now at step {current_step} and your current observation is: {current_observation}
Your admissible actions of the current situation are: [{admissible_actions}].

Now it's your turn to take an action.
You should first reason step-by-step about the current situation. This reasoning process MUST be enclosed within <think> </think> tags.
Once you've finished your reasoning, you should choose an admissible action for current step and present it within <action> </action> tags.
The special action 'call_cloud' delegates only the current step to a cloud model. After that single action, you resume control.
\end{lstlisting}

\par\medskip
\noindent\textbf{WebShop.}\par
\smallskip

\begin{lstlisting}
You are an expert autonomous agent operating in the WebShop e-commerce environment.
Your task is to: {task_description}.
Prior to this step, you have already taken {step_count} step(s). Below are the most recent {history_length} observations and the corresponding actions you took: {action_history}
You are now at step {current_step} and your current observation is: {current_observation}.
Your admissible actions of the current situation are:
[
{available_actions}
].

Now it's your turn to take one action for the current step.
You should first reason step-by-step about the current situation, then think carefully which admissible action best advances the shopping goal. This reasoning process MUST be enclosed within <think> </think> tags.
Your entire response MUST contain both the nonempty <think> block and exactly one <action> block, in that order.
Use this exact layout: <think>brief current-step reasoning</think><action>search[red shoes]</action>
The action must be either a search query or a click on one of the listed clickables. Do not use Markdown or add text outside the two tags.
\end{lstlisting}

When the on-device SLM selects \texttt{call\_cloud}, the cloud model receives
the complete interaction history and the current admissible action set. We use
the following benchmark-specific prompts to constrain it to one executable
action.

\par\medskip
\noindent\textbf{ALFWorld.}\par
\smallskip

\begin{lstlisting}
You are the one-step cloud controller for an ALFWorld episode.

Choose exactly one action for the current state.

Return only <action>ONE EXACT ADMISSIBLE ACTION</action>.

Do not return a plan, explanation, multiple actions, or call_cloud.

Task: {task}

Inventory: {inventory_or_empty}

Complete action-observation-result trajectory:

{full_action_observation_result_trajectory}

Current observation: {current_observation}

Current admissible actions:

- {admissible_action_1}
- {admissible_action_2}
...
\end{lstlisting}

\par\medskip
\noindent\textbf{WebShop.}\par
\smallskip

\begin{lstlisting}
You are the one-step cloud controller for a WebShop episode.

Choose exactly one action for the current state.

Return exactly one action using this format:

<action>ONE EXACT WEBSHOP ACTION</action>

Do not output <think>, reasoning, explanation, Markdown, multiple actions, or call_cloud.

The action must be either:

- search[nonempty query]

- click[one exact action from the available actions]

Shopping instruction: {task}

Complete action-observation-result trajectory:

{history}

Current observation: {current_observation}

Current available actions:

- {available_action_1}
- {available_action_2}
- {available_action_3}
...
\end{lstlisting}

\subsection{Trajectory--Skill Generation}
\label{app:skill-generation}

In Stage~1, the cloud trajectory teacher receives a task instruction
$x$, an SLM-generated trajectory $\tau^{\mathrm{S}}$, and its return
$R(\tau^{\mathrm{S}})$, and produces a JSON annotation containing an episode
summary and a reusable episode skill. Table~\ref{tab:skill-io} shows a real
ALFWorld example from the annotation data. The source trajectory contains 30
steps and ends in failure; we summarize the trajectory to fit the layout and
reproduce the generated fields verbatim.

\paragraph{Skill-generation prompt.}
The trajectory teacher is prompted with the following template. The
\texttt{episode\_skill\_instruction} is selected according to the episode
outcome.

\begin{lstlisting}
Analyze the following agent episode and return ONLY valid JSON.

You need to complete two fields:

1. Write a concise episode_summary.
2. {episode_skill_instruction}

Important constraints:

- Write episode_skill as one short policy-facing skill, not as retrospective
  explanation of the trajectory.
- Return only these top-level fields: episode_summary, episode_skill.

Return format:
{
  "episode_summary": "string",
  "episode_skill": "string"
}

Episode context:

- Task description: {task_description}
- episode_success: {success|failure|unknown}
- Interaction trajectory:

{formatted_trajectory}

For a successful trajectory, write one episode_skill that extracts the
successful trajectory into a workflow: the core decision rule and action
ordering that made the trajectory work.

For a failed trajectory, write one episode_skill that extracts the failed
trajectory into avoidance rules: the core mistake and warning signs that the
agent should avoid.

The formatted trajectory is:

Step {step_index}

Observation: {observation}

Response: {model_response}

Action valid: {true|false}
\end{lstlisting}

\begin{table*}[t]
    \centering
    \begin{minipage}[t]{0.48\textwidth}
        \vspace{0pt}
        \textbf{Input}
        \begin{lstlisting}
{
  "task_description":
    "put two candle in drawer.",
  "episode_success": "failure",
  "candidate_step_indices": [0, 1, ..., 29],
  "trajectory": [
    {
      "step_index": 0,
      "observation": "-= Welcome to TextWorld, ALFRED! =- ...
        Your task is to: put two candle in drawer.",
      "response": "<think>...</think><action>go to countertop 1</action>",
      "actions": ["go to countertop 1"],
      "action_valid": true
    },
    ...,
    {
      "step_index": 29,
      "observation": "Nothing happens.",
      "response": "<think>...</think><action>go to drawer 1</action>",
      "actions": ["go to drawer 1"],
      "action_valid": true
    }
  ]
}
        \end{lstlisting}
    \end{minipage}
    \hfill
    \begin{minipage}[t]{0.48\textwidth}
        \vspace{0pt}
        \textbf{Output}
        \begin{lstlisting}
{
  "episode_summary": "The agent failed to complete the task of putting two candles in a drawer. It systematically searched drawers and countertops but never located any candles. During the search, the agent exhibited inefficient behaviors such as repeatedly visiting already-checked empty drawers, picking up and immediately replacing irrelevant items (cloth), and taking unrelated objects (dishsponge) without purpose. The episode ended with the agent still searching and holding an irrelevant item, having made no progress toward the actual goal.",

  
  "episode_skill": "When searching for target objects, strictly avoid interacting with or picking up items that do not match the task description; if a location has been confirmed empty or containing only irrelevant items, do not revisit it, and maintain focus on unexplored areas until the target object is found."
}
        \end{lstlisting}
    \end{minipage}
    \caption{A real ALFWorld trajectory-to-skill example from Stage~1. The
    input preserves the original JSON fields and selected trajectory entries;
    ellipses indicate omitted intermediate steps. The output is reproduced
    from the corresponding annotation.}
    \label{tab:skill-io}
\end{table*}

\subsection{Models and Training}
We train Qwen3-0.6B or Qwen3-1.7B as the on-device SLMs. Qwen3.7-Max is the
primary cloud model, and Gemini 3.1 Pro is used for the separate cloud-model
comparison. The primary ALFWorld run uses 8 AMD Instinct MI308X GPUs.

Stage~1 first collects 180 training tasks with eight trajectories per task,
yielding 1,440 trajectory--skill examples (1,296 train and 144 validation).
Skill SFT uses three epochs, maximum sequence length 8,192, learning rate
$5\times10^{-6}$, and batch size 8. We then perform action/protocol-repair SFT
on 512 train and 64 validation examples for one epoch, with maximum sequence
length 4,096, learning rate $2\times10^{-6}$, and batch size 8. Consultation-free
RL runs for 160 updates at learning rate $10^{-6}$, with group size 8, batch
size 256, $\beta_{\mathrm{opd}}=5.0$, and
$\lambda_{\mathrm{opd}}=0.01$.

Stage~2 starts from the Stage~1 policy. Failure mining uses 1,024 label-training
games, 256 validation games, and 256 runtime games; each failed trajectory
contributes at most three candidate states, and replay uses the exact prefix.
The resulting cold-start SFT train for two epochs with maximum sequence
length 4,096, learning rate $2\times10^{-6}$, and batch size 16.

Stage~3 continues in the augmented action space, where each consultation
executes one cloud action before local control resumes. We train for one
epoch over these rows, yielding 160 optimizer updates, and use learning rate
$10^{-6}$, group size 8, batch size 256, $\alpha_{\mathrm{step}}=0.5$,
$\alpha_{\mathrm{cf}}=0.25$, and $\lambda_{\mathrm{opd}}=0.01$.

\section{Supplementary Results}
\label{app:results}

\subsection{Cross-Domain Generalization}
\label{app:cross-domain}

\begin{wraptable}{r}{0.45\textwidth}
    \centering
    \small
    \setlength{\tabcolsep}{6pt}
    \renewcommand{\arraystretch}{1.05}
    \caption{Unseen-split success by task type.}
    \label{tab:unseen-breakdown}
    \begin{tabular}{lc}
        \toprule
        Task type & Success (\%) \\
        \midrule
        Pick  & 91.7 \\
        Look  & 100.0 \\
        Clean & 100.0 \\
        Heat  & 95.7 \\
        Cool  & 95.2 \\
        Pick2 & 94.1 \\
        \midrule
        \textbf{Overall} & \textbf{96.3} \\
        \bottomrule
    \end{tabular}
\end{wraptable}
On the ALFWorld Unseen subset, we achieve 96.3\%, compared with 94.3\% on Seen, for a combined score of 95.3\%. The Unseen rooms and layouts were not observed during training, demonstrating strong out-of-distribution (OOD) generalization. Success is at least 94.1\% on five task types, with 100\% on Look and Clean (Table~\ref{tab:unseen-breakdown}). This strong OOD performance may be due to the way the model consults the cloud, which makes the impact of OOD on its performance small.

\subsection{Analysis by Task Type}
\label{app:task-analysis}

Table~\ref{tab:task-breakdown} compares success and average cloud calls per
episode for each task type. On Seen, Pick and Clean achieve the highest
success, whereas Heat and Pick2 are the most challenging and use the most
consultations. On Unseen, success improves for every task type except Pick:
Heat and Pick2 gain 8.15 and 6.62 percentage points, respectively, while
Look and Clean reach 100\%. The consultation pattern also changes across
splits. Pick, Look, and Clean use more calls on Unseen, whereas Cool, Heat,
and Pick2 use fewer. Thus, the checkpoint does not simply consult more on
tasks with lower success; its consultation demand varies with the task
structure and the encountered layout.

\begin{table*}[t]
    \centering
    \small
    \setlength{\tabcolsep}{8pt}
    \renewcommand{\arraystretch}{1.05}
    \caption{Task-wise success and average cloud calls per episode on the Seen
    and Unseen splits.}
    \label{tab:task-breakdown}
    \begin{tabular}{lrrrr}
        \toprule
        Task & Seen success (\%) & Seen calls/ep. & Unseen success (\%) & Unseen calls/ep. \\
        \midrule
        Pick  & 100.0 & 0.1 & 91.7 & 0.5 \\
        Look  & 92.3  & 0.5 & 100.0 & 0.7 \\
        Clean & 100.0 & 0.5 & 100.0 & 0.7 \\
        Cool  & 92.0  & 1.2 & 95.2 & 0.7 \\
        Heat  & 87.5  & 1.5 & 95.7 & 0.8 \\
        Pick2 & 87.5  & 1.3 & 94.1 & 0.6 \\
        \bottomrule
    \end{tabular}
\end{table*}

\section{Cases Study}
\label{app:rescue_cases}
Success rates quantify the overall benefit of consultation but do not establish whether the cloud model’s input was strictly necessary. To examine this question, we analyze four representative rescue cases from the evaluation traces in Table 7 under aligned comparison settings. Both ALFWorld branches use the same 30-step budget, while both WebShop branches use the same 15-step budget, start from the same restored session, and share the same frozen edge continuation, differing only in the action taken at the disputed state. We present two representative cases from each benchmark to illustrate our central finding: targeted cloud consultation can resolve decision bottlenecks and recover successful trajectories.

In all four cases, the on-device SLM makes an incorrect decision when operating alone. The cloud model is consulted at this decision bottleneck and provides an executable corrective action; the on-device SLM then resumes control and completes the task successfully. This reflects Sibyl’s intended division of labor: the cloud model resolves a difficult local decision, while the edge model executes the remaining routine steps.

The two ALFWorld cases involve cloud-guided search and target disambiguation. Once the cloud model identifies the correct location or target, the on-device SLM correctly performs subsequent actions such as picking up, cleaning, and placing the object. The two WebShop cases are more localized: the on-device SLM fails to select the required product option on its own, whereas the cloud model selects the correct option. The on-device SLM then executes the same final “buy now” action, which succeeds with the correct option in place. Together, these traces provide qualitative evidence that Sibyl invokes the cloud at decision bottlenecks, corrects local edge-model errors, and delegates the remaining workflow to the on-device model.
\begin{table*}[t]
    \centering
    \small
    \setlength{\tabcolsep}{4pt}
    \renewcommand{\arraystretch}{1.15}
    \begin{tabularx}{\textwidth}{@{}>{\raggedright\arraybackslash}p{0.18\textwidth}
        >{\raggedright\arraybackslash}p{0.23\textwidth}
        >{\raggedright\arraybackslash}p{0.25\textwidth}X@{}}
        \toprule
        \textbf{Case} & \textbf{Edge-only branch} &
        \textbf{Cloud's critical action} & \textbf{Edge continuation and outcome} \\
        \midrule
        \textbf{ALFWorld: clean butterknife} &
        30 steps, no cloud calls, and failure. The on-device SLMs takes an
        ordinary \texttt{knife}, cleans and returns it, and never reaches the
        target \texttt{butterknife}. &
        The cloud steers the search to \texttt{diningtable 3} at step 20,
        where \texttt{butterknife 1} is visible. &
        The on-device SLMs takes the butterknife, cleans it at the sink, and moves
        it to \texttt{sidetable 1}. The collaborative run succeeds in 25 steps
        with 3 cloud calls. \\

        \addlinespace

        \textbf{ALFWorld: two credit cards} &
        30 steps, no cloud calls, and failure while searching for the second
        card after placing the first. &
        The cloud first requests \texttt{inventory}, then chooses
        \texttt{diningtable 1} to locate \texttt{creditcard 1}, and finally
        \texttt{armchair 1} as the destination. &
        The on-device SLMs takes the card and executes the final move locally.
        The collaborative run succeeds in 14 steps with 3 cloud calls. \\

        \addlinespace

        \textbf{WebShop: random-color bath mitt} &
        From the same product page, the forced edge action is
        \texttt{click[buy now]}; the purchase has empty options and scores
        $0.75$. &
        The cloud selects the required option
        \texttt{click[random color 1pc]}. &
        The identical edge continuation then executes
        \texttt{click[buy now]}, producing the correct option and score
        $1.00$. \\

        \addlinespace

        \textbf{WebShop: ottoman size} &
        From the same product page, the forced edge action is
        \texttt{click[buy now]}; the purchase has empty options and scores
        $0.80$. &
        The cloud selects the required size
        \texttt{click[100x40x40cm]}. &
        The identical edge continuation executes
        \texttt{click[buy now]}, producing the correct size and score
        $1.00$. \\
        \bottomrule
    \end{tabularx}
    \caption{Representative edge–cloud rescue cases. The cloud provides the bottleneck action shown in the third column; subsequent actions are executed by the on-device SLM. The two benchmarks follow their respective evaluation protocols.}
    \label{tab:rescue_cases}
\end{table*}

\section{Limitations and Future Work}
\label{app:limitations}

\method{} focuses on when an on-device agent should request one-step assistance from a fixed cloud model. This formulation assumes that the cloud service is available when needed and that the agent can share sufficient task context to obtain a useful action. Privacy-sensitive applications and intermittent connectivity introduce additional constraints: the agent may need to limit the context it transmits or continue safely when a consultation fails. Future work could incorporate privacy-aware context selection and fallback behavior into the consultation policy.

The current formulation also treats the cloud as a single source of assistance. A deployed agent may instead have access to models with different capabilities, prices, and response times. Extending \method{} to choose both when to consult and which model to consult could improve its flexibility under changing workloads and service conditions, while retaining local execution as the default.

\end{document}